\documentclass[pdflatex,iicol,sn-mathphys-num]{sn-jnl}

\usepackage{graphicx}%
\usepackage{multirow}%
\usepackage{amsmath,amssymb,amsfonts}%
\usepackage{amsthm}%
\usepackage{mathrsfs}%
\usepackage[title]{appendix}%
\usepackage{xcolor}%
\usepackage{textcomp}%
\usepackage{manyfoot}%
\usepackage{booktabs}%
\usepackage{algorithm}%
\usepackage{algorithmicx}%
\usepackage{algpseudocode}%
\usepackage{listings}%
\usepackage{tabularx}
\usepackage{array}
\usepackage{subcaption}
\usepackage{rotating}
\usepackage{makecell}
\usepackage{placeins}
\usepackage{textgreek}

\theoremstyle{thmstyleone}

\theoremstyle{thmstyletwo}%

\theoremstyle{thmstylethree}%

\begin{document}

\title[CSCWD for Tiny Object Detection on Edge Devices]
{CSCWD: Cross-Scale Channel-wise Knowledge Distillation for
	Lightweight Tiny Object Detection on Edge Devices}

\author[1]{\fnm{Amir} \sur{Zamani}}
\email{amirzamani@cuir.ac.ir}

\author*[2]{\fnm{Zeinab} \sur{Ghasemi-Naraghi}}
\email{naraghi@kntu.ac.ir}

\affil[1]{%
	\orgdiv{Department of Computer Engineering},
	\orgname{Islamic Revolution Comprehensive University},
	\orgaddress{
		\city{Tehran},
		\country{Iran}}
}

\affil*[2]{%
	\orgdiv{Department of Computer Engineering},
	\orgname{K.N. Toosi University of Technology},
	\orgaddress{
		\city{Tehran},
		\country{Iran}}
}

\abstract{
	Real-time tiny object detection in aerial imagery is constrained by the weak
	spatial evidence of very small objects and the loss of high-resolution detail
	in lightweight detectors. This study presents Cross-Scale Channel-wise
	Knowledge Distillation (CSCWD), a training-time framework that transfers
	high-resolution spatial representations from a YOLO11m-P2 teacher to a compact
	YOLO11n student without altering the student's inference architecture. Unlike
	conventional same-scale feature distillation, CSCWD transfers supervision from
	teacher P\textsubscript{2} to student P\textsubscript{3} after feature alignment
	while retaining same-scale distillation at deeper pyramid levels. Under the
	unified seven-sequence Drone-vs-Bird validation protocol, YOLO11n-CSCWD
	achieves 50.17\% mean average precision at an intersection-over-union threshold
	of 0.5 (mAP@0.5) and 59.73\% recall, improving the matched CA-YOLO11n baseline
	by 2.92 percentage points in mAP@0.5 and 3.55 points in recall. Cross-scale
	alignment further increases mAP@0.5 by 2.09 points over the corresponding
	same-scale channel-wise distillation configuration. In zero-shot evaluation on
	DUT-Anti-UAV, mAP@0.5 increases from 48.29\% to 50.06\% without target-domain
	fine-tuning. This domain was included because its challenging small targets make
	low-latency, computationally efficient detection particularly relevant. On
	Raspberry Pi~5 using NCNN-FP16 at 640$\times$640 resolution, the
	2.58-million-parameter student achieves 50.32\% mAP@0.5 at 82.32\,ms mean
	wall-clock latency, or 12.15 frames per second, while retaining essentially the
	same runtime and memory requirements as the matched baseline. The results
	support cross-scale distillation for improving tiny-target detection without
	increasing inference-time model complexity.
}


\maketitle

\section{Introduction}\label{sec1}

Tiny object detection is difficult because targets spanning only a few pixels
provide limited appearance information and weak localization evidence
\cite{nikouei_small_2025}. In long-range visual detection of small objects,
problems such as imaging distance, scale variation, motion blur, low contrast,
and background clutter further weaken these cues. Long-range visual detection
of unmanned aerial vehicles (UAVs) is a representative instance of this
problem because targets frequently occupy tiny image regions while practical
detectors operate under latency, memory, and computational constraints on
embedded hardware
\cite{coluccia_drone-vs-bird_2024,zhou_improved_2025}. Accordingly, this study
evaluates the proposed approach on UAV imagery, where preserving fine spatial
evidence under real-time edge constraints is relevant.

Existing approaches improve tiny-target representation through multi-scale
processing, higher-capacity backbones, and temporal compensation.
Representative Drone-vs-Bird methods use multi-scale inference, wide residual
networks, or tracking-assisted recovery of missed detections
\cite{laroca_improving_2025,wong_wrn-yolo_2025,sawada_tiny_2025}. These
strategies can strengthen weak-target representation, but their additional
computation and memory demand can increase inference latency on low-power
platforms \cite{mittal_comprehensive_2024}. Lightweight detectors reduce
inference complexity, yet the accompanying loss of feature resolution and
representational capacity can discard shallow high-resolution cues needed for
tiny-target localization. This accuracy--efficiency tension motivates
training-time mechanisms that enrich a compact detector without enlarging its
network.

Knowledge distillation (KD) transfers information from a high-capacity teacher
to a compact student during training
\cite{mansourian_comprehensive_2025}. Feature-level KD is particularly relevant
to detection because intermediate maps retain spatial structure that final
predictions do not preserve. Channel-wise knowledge distillation (CWD)
matches channel-wise spatial distributions, but conventional scale-matched
alignment cannot directly exploit a high-resolution teacher feature level when
the student has no corresponding pyramid level
\cite{shu_channel-wise_2021,zhu_scalekd_2023}. For tiny objects, this mismatch
restricts the transfer of localization-sensitive spatial detail, motivating
cross-scale feature distillation.

\begin{figure}[t!]
	\centering
	\includegraphics[width=\columnwidth]{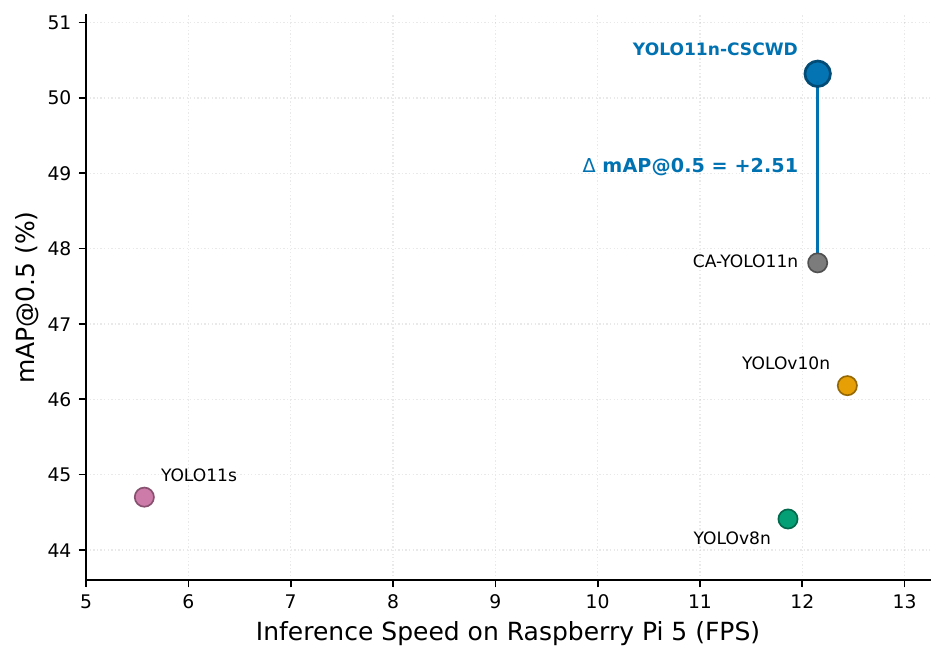}
	\caption{
		Deployed accuracy and inference speed of lightweight detectors on
		Raspberry Pi~5 using NCNN-FP16 at $640\times640$ under the unified
		seven-sequence validation protocol
	}
	\label{fig:fig1}
\end{figure}

This work presents Cross-Scale Channel-wise Knowledge Distillation (CSCWD)
using a YOLO11m-P2 teacher with a high-resolution $P_2$ detection branch
and a standard YOLO11n student \cite{ultralytics_yolov11_2024}. CSCWD transfers
supervision from teacher $P_2$ to student $P_3$ while retaining matched-scale
distillation at $P_4$ and $P_5$. Teacher and auxiliary distillation components
are removed after training, leaving the student inference graph unchanged.
The resulting student is evaluated against
CA-YOLO11n~\cite{zamani_optimizing_2026} and lightweight detectors on
Raspberry Pi~5; Fig.~\ref{fig:fig1} summarizes deployed accuracy and inference
speed.

The main contributions are summarized as follows:
\begin{itemize}
	\item A cross-scale CWD formulation transfers high-resolution teacher
	knowledge from $P_2$ to student $P_3$ while retaining deeper same-scale
	distillation, targeting the loss of fine spatial information in lightweight
	tiny-object detection.
	
	\item A deployment-preserving teacher--student formulation restricts
	high-resolution supervision to training, allowing the distilled model to
	retain the original YOLO11n inference graph and computational footprint.
	
	\item Controlled ablations, Drone-vs-Bird evaluation, zero-shot
	DUT-Anti-UAV testing, and physical Raspberry Pi~5 deployment quantify
	detection gains, zero-shot cross-dataset performance, and embedded inference
	efficiency
	\cite{coluccia_drone-vs-bird_2025,zhao_vision-based_2022}.
\end{itemize}

\section{Related Work}
\label{sec:related}

Prior work relevant to lightweight tiny-object detection has sought to
strengthen weak-target representation through multi-scale inference, increased
feature-extraction capacity, and temporal compensation. On Drone-vs-Bird,
Laroca et al. combine multi-scale processing with data augmentation and temporal
post-processing, Wong et al. employ a WideResNet-based detector, and Sawada
et al. recover missed detections using tracking-assisted temporal information
\cite{laroca_improving_2025,wong_wrn-yolo_2025,sawada_tiny_2025}.
These approaches improve detection performance through different mechanisms
but add computation or pipeline complexity, limiting their suitability when
deployment complexity must remain low. Context-aware augmentation offers a
training-only alternative that improves a nano-scale detector without modifying
its inference graph \cite{zamani_optimizing_2026}; this augmentation strategy,
however, does not directly provide teacher-derived feature supervision.

Building on the original KD formulation \cite{hinton_distilling_2015},
feature-level methods include intermediate representation regression, attention
transfer, masked generative distillation, and focal-and-global knowledge
distillation
\cite{romero_fitnets_2015,zagoruyko_paying_2017,
	avidan_masked_2022,yang_focal_2022}. These methods differ in the form of
supervision, but they share the goal of improving the student through
training-time representation transfer without adding inference modules. CWD
represents feature knowledge through channel-wise spatial probability
distributions, making channel-specific spatial structure explicit during
transfer \cite{shu_channel-wise_2021}. Many feature-level formulations still
rely on matching selected teacher and student representations at compatible
or corresponding scales. ScaleKD specifically addresses scale-aware
distillation for small-object detection through a scale-decoupled feature
distillation module and a cross-scale assistant
\cite{zhu_scalekd_2023}. A narrower architectural gap arises when the teacher
contains a high-resolution $P_2$ level while a lightweight student begins at
$P_3$, leaving no direct same-scale counterpart for teacher $P_2$. Transferring
this teacher-only high-resolution representation while preserving the
student's original inference graph motivates the proposed cross-scale
$P_2$-to-$P_3$ transfer.

\section{Proposed Method}
\label{sec:method}

The proposed CSCWD framework transfers teacher-only high-resolution
representations to a lightweight student during training while preserving the
student's original inference architecture.

\subsection{Framework Overview}
\label{sec:overview}

\begin{figure*}[!t]
	\centering
	\includegraphics[width=0.95\textwidth]{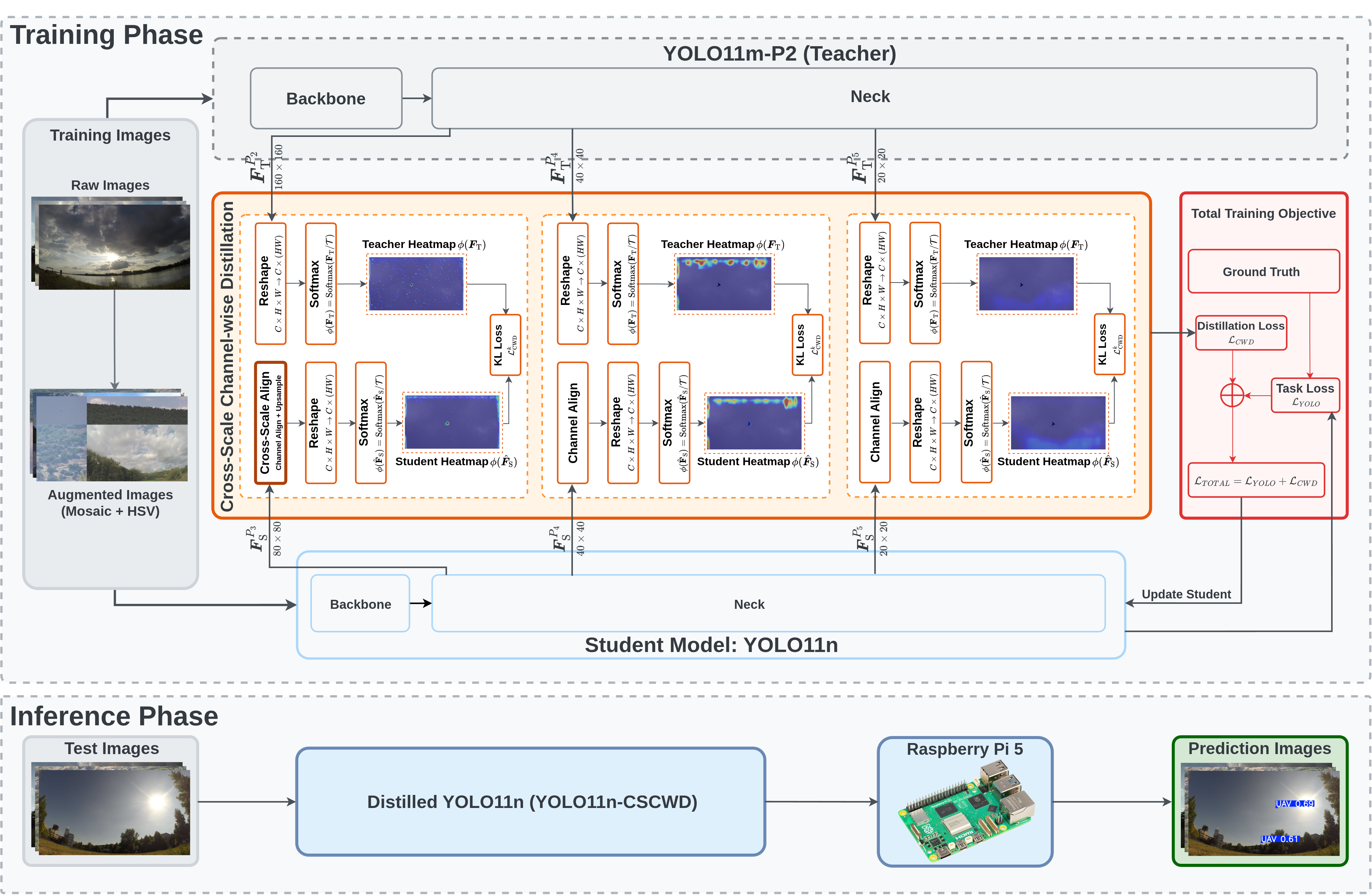}
\caption{
	Overview of CSCWD training and inference. The frozen YOLO11m-P2
	teacher provides cross-scale $P_2$-to-$P_3$ supervision and matched-scale
	distillation at $P_4$ and $P_5$ during training. Inference retains only the
	distilled YOLO11n-CSCWD model. Raspberry Pi image: CC BY 2.0~\cite{sparkfun_raspberrypi5_image}
}
	\label{fig:framework}
\end{figure*}

CSCWD uses an asymmetric teacher--student configuration to transfer
high-resolution supervision that is unavailable in the student's native
feature pyramid. During training, the frozen YOLO11m-P2 teacher provides
cross-scale $P_2$-to-$P_3$ supervision together with matched-scale
distillation at $P_4$ and $P_5$, while the YOLO11n student remains the only
network retained for deployment. Figure~\ref{fig:framework} summarizes the
complete training and inference flow.

\subsection{Asymmetric Teacher--Student Design}
\label{sec:asymmetric}

The teacher and student are deliberately asymmetric in model capacity and
feature resolution. The teacher extends YOLO11m
\cite{ultralytics_yolov11_2024} with a high-resolution $P_2$ detection
branch, producing a $160\times160$ feature map for a $640\times640$ input.
The teacher is trained in two stages: VisDrone~\cite{zhu_detection_2022}
adaptation followed by target-domain fine-tuning. The corresponding optimization and augmentation
settings are summarized in Table~\ref{tab:augmentation}.

\begin{table}[b]
	\centering
	\caption{Teacher training settings for VisDrone adaptation and target fine-tuning.}
	\label{tab:augmentation}
	
	\footnotesize
	\setlength{\tabcolsep}{2pt}
	\renewcommand{\arraystretch}{1.05}
	
	\begin{tabularx}{\columnwidth}{
			@{}
			>{\raggedright\arraybackslash}p{0.27\columnwidth}
			>{\centering\arraybackslash}p{0.34\columnwidth}
			>{\centering\arraybackslash}p{0.34\columnwidth}
			@{}
		}
		\toprule
		\textbf{Setting}
		& \textbf{Stage~1: VisDrone adaptation}
		& \textbf{Stage~2: Target fine-tuning} \\
		\midrule
		
		Initialization
		& COCO (\texttt{yolo11m.pt})
		& Stage-1 best \\
		
		Epochs / batch
		& 30 / 4
		& 100 / 12 \\
		
		Optimizer
		& Default
		& SGD \\
		
		Mosaic ($p$/close)
		& 1.0 / last 5
		& 0.75 / last 10 \\
		
		MixUp
		& 0.0
		& 0.15 \\
		
		Scale / translate
		& 0.5 / 0.1
		& 0.5 / 0.0 \\
		
		Flip (V/H)
		& 0.5 / 0.5
		& 0.0 / 0.0 \\
		
		HSV (H/S/V)
		& 0.015 / 0.7 / 0.4
		& 0.0 / 0.0 / 0.4 \\
		
		\bottomrule
	\end{tabularx}
\end{table}

The student retains the standard YOLO11n architecture without a $P_2$
branch and therefore begins its detection pyramid at $P_3$. It is initialized
from COCO-pretrained weights and adapted on VisDrone before CSCWD training.
This asymmetry preserves the lightweight student graph while exposing it
during training to spatial information available only at the teacher's
higher-resolution $P_2$ level.

\subsection{Cross-Scale Feature Alignment}
\label{sec:alignment}

CSCWD transfers supervision from teacher $P_2$ to student $P_3$, while
$P_4$ and $P_5$ are distilled at matched scales. For loss computation, the
student representations are aligned to the corresponding teacher feature
spaces:
\begin{equation}
S{:}P_3 \rightarrow T{:}P_2,\qquad
S{:}P_4 \rightarrow T{:}P_4,\qquad
S{:}P_5 \rightarrow T{:}P_5
\end{equation}
The first pair defines the cross-scale transfer, whereas the remaining two
pairs are scale matched. For loss computation, each student representation is
projected to the channel dimension of its paired teacher feature. The
$S{:}P_3$ feature is additionally upsampled to the spatial resolution of
$T{:}P_2$; $P_4$ and $P_5$ require channel projection only.

\subsection{Cross-Scale Channel-wise Knowledge Distillation}
\label{sec:cwd_formulation}

To transfer the aligned feature representations, the framework extends
Channel-wise Distillation (CWD)~\cite{shu_channel-wise_2021} to support
cross-scale matching. Unlike feature-regression approaches such as
FitNets~\cite{romero_fitnets_2015}, CWD converts each feature channel into a
spatial probability distribution and minimizes the Kullback--Leibler (KL)
divergence between the teacher and student distributions.

Let $T$ and $S$ denote the frozen teacher and trainable student networks, respectively. For each feature pair defined by these alignment mappings
$P_3 \rightarrow P_2$, $P_4 \rightarrow P_4$, and $P_5 \rightarrow P_5$,
the intermediate feature representations are denoted as:

\begin{equation}
	\mathbf{F}_{\mathrm{T}}^{k} \in
	\mathbb{R}^{N \times C_{\mathrm{T}}^{k} \times H_{\mathrm{T}}^{k} \times W_{\mathrm{T}}^{k}}
\end{equation}

and

\begin{equation}
	\mathbf{F}_{\mathrm{S}}^{k} \in
	\mathbb{R}^{N \times C_{\mathrm{S}}^{k} \times H_{\mathrm{S}}^{k} \times W_{\mathrm{S}}^{k}},
\end{equation}

where $N$ represents the batch size, and $C$, $H$, and $W$ denote the channel, height, and width dimensions of the feature tensors. For the cross-scale pair ($P_3\rightarrow P_2$), $\mathbf{F}_{\mathrm{T}}^{k}$ specifically represents the teacher's high-resolution $P_2$ feature.

\subsubsection*{Channel alignment and spatial matching}

Since teacher and student features have different channel dimensions, the student representation is first projected into the teacher feature space using a learnable $1\times1$ convolution. The channel transformation function is defined as:

\begin{equation}
	g_{\mathrm{map}}^{k} :
	\mathbb{R}^{N \times C_{\mathrm{S}}^{k} \times H_{\mathrm{S}}^{k} \times W_{\mathrm{S}}^{k}}
	\longrightarrow
	\mathbb{R}^{N \times C_{\mathrm{T}}^{k} \times H_{\mathrm{S}}^{k} \times W_{\mathrm{S}}^{k}}.
\end{equation}

The channel-aligned student feature is obtained by:

\begin{equation}
	\tilde{\mathbf{F}}_{\mathrm{S}}^{k}
	=
	g_{\mathrm{map}}^{k}
	\big(
	\mathbf{F}_{\mathrm{S}}^{k}
	\big).
\end{equation}

If the spatial resolution does not match the corresponding teacher feature, bilinear interpolation is applied:

\begin{equation}
	\mathrm{Upsample} :
	\mathbb{R}^{N \times C_{\mathrm{T}}^{k}
		\times H_{\mathrm{S}}^{k}
		\times W_{\mathrm{S}}^{k}}
	\longrightarrow
	\mathbb{R}^{N \times C_{\mathrm{T}}^{k}
		\times H_{\mathrm{T}}^{k}
		\times W_{\mathrm{T}}^{k}}.
\end{equation}

Therefore, the final aligned student feature is obtained as:

\begin{equation}
	\resizebox{0.98\columnwidth}{!}{$
		\displaystyle
		\hat{\mathbf{F}}_{\mathrm{S}}^{k}
		=
		\begin{cases}
			\mathrm{Upsample}
			\left(
			\tilde{\mathbf{F}}_{\mathrm{S}}^{k};
			H_{\mathrm{T}}^{k},W_{\mathrm{T}}^{k}
			\right),
			&
			\text{if }
			(H_{\mathrm{S}}^{k},W_{\mathrm{S}}^{k})
			\neq
			(H_{\mathrm{T}}^{k},W_{\mathrm{T}}^{k}),
			\\[2pt]
			\tilde{\mathbf{F}}_{\mathrm{S}}^{k},
			&
			\text{otherwise}.
		\end{cases}
		$}
\end{equation}

After alignment, teacher and student features share identical dimensions and can be compared through the distillation objective.

\subsubsection*{Channel-wise probability distributions}

Following the CWD formulation, each feature channel is normalized independently over spatial positions using a temperature-scaled softmax. Let
$x_{c,i}^{k}$ denote the activation value of channel $c$ at spatial position $i$. The spatial probability distribution is defined as:

\begin{equation}
	\phi\big(x_{c,i}^{k}\big)
	=
	\frac{\exp\!\left(x_{c,i}^{k}/\mathcal{T}\right)}
	{\sum_{j=1}^{H^{k}W^{k}}
		\exp\!\left(x_{c,j}^{k}/\mathcal{T}\right)},
	\label{eq:cwd_softmax}
\end{equation}

where $\mathcal{T}$ denotes the temperature parameter. This normalization converts each channel activation map into a spatial probability distribution, allowing the student to learn the teacher's spatial attention patterns.

\subsubsection*{Channel-wise distillation loss}

The distillation loss is calculated by minimizing the KL divergence between the teacher and aligned student channel distributions:

\begin{equation}
	\resizebox{0.98\columnwidth}{!}{$
		\displaystyle
		\mathcal{L}_{\mathrm{CWD}}^{k}
		=
		\frac{\mathcal{T}^{2}}{C^{k}}
		\sum_{c=1}^{C^{k}}
		\sum_{i=1}^{H^{k}W^{k}}
		\phi\!\left(\mathbf{F}_{\mathrm{T},c,i}^{k}\right)
		\log
		\left[
		\frac{
			\phi\!\left(\mathbf{F}_{\mathrm{T},c,i}^{k}\right)
		}{
			\phi\!\left(\hat{\mathbf{F}}_{\mathrm{S},c,i}^{k}\right)
		}
		\right]
		$}
	\label{eq:cwd_loss_single}
\end{equation}

where $C^{k}$ represents the number of channels at the corresponding aligned feature level. The complete distillation objective is obtained by aggregating all feature mappings:

\begin{equation}
	\mathcal{L}_{\mathrm{CWD}}
	=
	\alpha_{\mathrm{CWD}}
	\sum_{k\in\{P_3,P_4,P_5\}}
	\mathcal{L}_{\mathrm{CWD}}^{k},
	\label{eq:cwd_loss_multi}
\end{equation}

where $\alpha_{\mathrm{CWD}}$ controls the contribution of the distillation term.

\subsubsection*{Total training objective}

The final optimization objective combines the original detection loss and the proposed distillation loss:

\begin{equation}
	\mathcal{L}_{\mathrm{TOTAL}}
	=
	\mathcal{L}_{\mathrm{YOLO}}
	+
	\mathcal{L}_{\mathrm{CWD}}.
	\label{eq:total_loss}
\end{equation}

All teacher, projection, and spatial-alignment components are discarded after
training. Deployment therefore retains only the distilled YOLO11n student
(YOLO11n-CSCWD), preserving the original inference graph and computational
footprint.

\section{Experiments and Results}
\label{sec:experiments}

Experiments evaluate CSCWD under a unified Drone-vs-Bird protocol, zero-shot
cross-dataset testing, controlled ablations, and physical deployment on
Raspberry Pi~5.

\subsection{Experimental Setup and Evaluation Protocol}
\label{sec:experimental_protocol}

\textbf{Datasets and evaluation.}
Drone-vs-Bird~\cite{coluccia_drone-vs-bird_2025,coluccia_drone-vs-bird_2024}
is the primary benchmark. Frames are sampled at 2~FPS, and the seven-sequence
validation protocol of Laroca et al.~\cite{laroca_improving_2025} contains
12,381 labeled frames. The seven sequences form the validation split; the
reported metrics are computed separately for each sequence and averaged with
equal weight, while the remaining 70 of 77 sequences are used for target-domain
training. Zero-shot evaluation uses the 2,200-image DUT-Anti-UAV test
split~\cite{zhao_vision-based_2022}, without target-domain fine-tuning.

\textbf{Training configuration.}
CA-YOLO11n denotes the matched baseline trained with the context-aware
augmentation strategy of Zamani et al.~\cite{zamani_optimizing_2026}.
Before distillation, the student is initialized from COCO-pretrained YOLO11n
weights and adapted to VisDrone~\cite{zhu_detection_2022} for 50 epochs.
This aerial-domain adaptation uses Mosaic with probability 1.0, vertical and
horizontal flips with probability 0.5, and HSV color augmentation. CSCWD
training then proceeds for 100 epochs on a single NVIDIA RTX~5070~Ti GPU
with 16~GB VRAM using SGD (batch size 32, momentum 0.937, weight decay
$5\times10^{-4}$), unless a specific ablation variant is stated otherwise. The final distillation configuration uses $\mathcal{T}=4$ and
$\alpha_{\mathrm{CWD}}=150$.

\textbf{Metrics and deployment protocol.}
Detection performance is reported using Precision, Recall, mAP@0.5, and
mAP@0.5:0.95. Desktop PyTorch and deployed NCNN-FP16 accuracies are reported
separately. For intermediate-feature analysis, a normalized Target
Concentration Ratio (nTCR) is defined as a diagnostic of spatial selectivity.
Here, $z_c(u)$ denotes the activation of channel $c$ at spatial position $u$,
$C$ is the channel count, $H$ and $W$ are feature-map dimensions, $v$ spans
all spatial positions, $\Omega_{\mathrm{GT}}$ denotes the ground-truth region,
and $q(u)$ is the channel-averaged spatial probability.
\begin{equation}
	\resizebox{0.98\columnwidth}{!}{$
		\displaystyle
		\mathrm{nTCR}
		=
		\frac{\sum_{u\in\Omega_{\mathrm{GT}}} q(u)}
		{|\Omega_{\mathrm{GT}}|/(HW)},
		\qquad
		q(u)
		=
		\frac{1}{C}\sum_{c=1}^{C}
		\frac{\exp(z_c(u)/\mathcal{T})}
		{\sum_v \exp(z_c(v)/\mathcal{T})}
		$}
	\label{eq:ntcr_compact}
\end{equation}
Values above 1 indicate target-region enrichment relative to a spatially
uniform allocation. The matched comparison uses CA-YOLO11n and
YOLO11n-CSCWD at $P_3$ over the 11,128 validation frames containing at least
one ground-truth target; teacher $P_2$ is descriptive only because of its
different spatial resolution.

Raspberry Pi~5 measurements use the 8~GB model, a four-core 2.4~GHz ARM CPU,
Debian~13, NCNN-FP16, batch size~1, and the \texttt{performance} governor.
CA-YOLO11n and YOLO11n-CSCWD use a 10-image warm-up and three full
validation passes at 640 and 960; 1280 is accuracy-only.
Figure~\ref{fig:fig1} reference-detector FPS values use one full 640 pass on
the same hardware and backend. Mean wall-clock latency, 95th-percentile (P95) latency, FPS,
and peak resident set size (RSS) are reported.

\subsection{Benchmark Comparison on Drone-vs-Bird}
\label{sec:sota}

Controlled comparisons examine whether the CSCWD gain is retained against
representative lightweight YOLO detectors under the same Drone-vs-Bird
training and evaluation setting. YOLO11n-CSCWD is compared with YOLOv8n,
YOLOv10n, YOLO11s, the matched CA-YOLO11n baseline
\cite{zamani_optimizing_2026}, and the YOLO11m-P2 teacher. The controlled
detectors use family-specific VisDrone-adapted initialization and otherwise
follow the same Drone-vs-Bird training configuration. Table~\ref{tab:sota_comparison}
summarizes these controlled results together with contextual values reported
in the literature.

\begin{table*}[t]
	\centering
\caption{
	Comparison on Drone-vs-Bird. The first three rows report literature values
	obtained under different data splits or inference procedures; the remaining
	models are evaluated under the unified seven-sequence validation protocol.
}
	\label{tab:sota_comparison}
	
	\footnotesize
	\setlength{\tabcolsep}{2.6pt}
	\renewcommand{\arraystretch}{1.06}
	
	\begin{tabularx}{\textwidth}{
			@{}
			>{\raggedright\arraybackslash}p{0.19\textwidth}
			>{\raggedright\arraybackslash}X
			>{\centering\arraybackslash}p{0.085\textwidth}
			c c c c
			@{}
		}
		\toprule
		
		\textbf{Method}
		& \textbf{Architecture}
		& \textbf{\makecell{Split\\type}}
		& \textbf{\makecell{Params\\(M)}}
		& \textbf{GFLOPs}
		& \textbf{\makecell{mAP\\@0.5}}
		& \textbf{\makecell{mAP\\@0.5:0.95}}
		\\
		\midrule

		Wong et al.~\cite{wong_wrn-yolo_2025}
		& WideResNet-based
		& Frame-based
		& 100.70
		& 260.0
		& 75.70\textsuperscript{b}
		& -- \\
		
		Sawada et al.~\cite{sawada_tiny_2025}
		& Faster R-CNN + RFLA
		& Video-based
		& 23.50
		& 130.0
		& 64.18\textsuperscript{a}
		& -- \\
		
		Laroca et al.~\cite{laroca_improving_2025}
		& YOLO11m (whole image)
		& Video-based
		& 20.10
		& 68.0
		& 55.23\textsuperscript{c}
		& -- \\

		\midrule
		
		YOLOv8n~\cite{ultralytics_yolov8_2023}
		& YOLOv8n
		& Video-based
		& 3.20
		& 8.7
		& 45.74
		& 15.90 \\
		
		YOLOv10n~\cite{wang_yolov10_2024}
		& YOLOv10n
		& Video-based
		& 2.30
		& 6.7
		& 44.49
		& 15.42 \\
		
		YOLO11s~\cite{ultralytics_yolov11_2024}
		& YOLO11s
		& Video-based
		& 9.40
		& 21.6
		& 46.17
		& 16.37 \\
		
		CA-YOLO11n~\cite{zamani_optimizing_2026}
		& YOLO11n
		& Video-based
		& 2.58
		& 6.4
		& 47.25
		& 16.92 \\
		
		YOLO11n-CSCWD
		& YOLO11n
		& Video-based
		& 2.58
		& 6.4
		& \textbf{50.17}
		& \textbf{17.30} \\
		
		YOLO11m-P2
		& YOLO11m + P2 branch
		& Video-based
		& 20.82
		& 134.5
		& 56.75
		& 21.26 \\
		
		\bottomrule
	\end{tabularx}
	
	\vspace{0.25em}
	\begin{minipage}{\textwidth}
		\scriptsize
		\textsuperscript{a}Different seven-video split in the original study.
		\quad
		\textsuperscript{b}Intra-video frame-based random split.
		\quad
		\textsuperscript{c}Whole-image-only result; the complete multi-scale
		and temporal pipeline reports 73.90\%.
	\end{minipage}
\end{table*}

Under the unified seven-sequence validation protocol, YOLO11n-CSCWD
improves CA-YOLO11n by 2.92 percentage points in mAP@0.5 and achieves
4.00 points higher mAP@0.5 than the larger YOLO11s model, while
retaining the 2.58~M-parameter, 6.4-GFLOP YOLO11n inference graph.
YOLO11m-P2 reaches 56.75\% mAP@0.5 but requires 20.82~M parameters
and 134.5~GFLOPs; this higher-capacity teacher is used only during training.
Literature-reported values are treated as contextual references rather than
direct rankings because their data splits and inference procedures differ.

\subsection{Sequence-Level and Cross-Dataset Analysis}
\label{sec:scenario_generalization}

Sequence-level results show that YOLO11n-CSCWD outperforms CA-YOLO11n
on six of the seven sequences, with the largest gains on Swarm
($+10.34$ points) and Hillside ($+9.23$ points). GoPro~002 is the only
sequence showing a decrease, from 28.86\% to 23.48\% mAP@0.5,
indicating scenario-dependent behavior (Table~\ref{tab:sequence_results}).

\begin{table}[b]
	\centering
	\caption{Sequence-level mAP@0.5 under the unified seven-sequence
		validation protocol.}
	\label{tab:sequence_results}
	
	\footnotesize
	\setlength{\tabcolsep}{2.3pt}
	\renewcommand{\arraystretch}{1.00}
	
	\begin{tabularx}{\columnwidth}{
			@{}
			>{\raggedright\arraybackslash}X
			>{\centering\arraybackslash}c
			>{\centering\arraybackslash}c
			>{\centering\arraybackslash}c
			@{}
		}
		\toprule
		
		\textbf{Sequence}
		& \textbf{\makecell{CA-\\YOLO11n}}
		& \textbf{\makecell{YOLO11n-\\CSCWD}}
		& \textbf{\makecell{YOLO11m-\\P2}} \\
		\midrule
		
		Mavic       & 91.28 & 91.41 & 93.07 \\
		Inspire     & 31.86 & 35.99 & 36.56 \\
		Disco       & 79.33 & 79.48 & 84.38 \\
		GOPR5843    & 26.80 & 28.63 & 37.73 \\
		Swarm       & 48.98 & 59.32 & 56.65 \\
		Hillside    & 23.61 & 32.84 & 54.96 \\
		GoPro~002   & 28.86 & 23.48 & 33.87 \\
		
		\bottomrule
	\end{tabularx}
\end{table}

\begin{figure*}[t]
	\centering
	
	\begin{minipage}[b]{0.235\textwidth}
		\centering
		\textbf{Ground Truth}\\[0.5mm]
		\includegraphics[width=\linewidth]{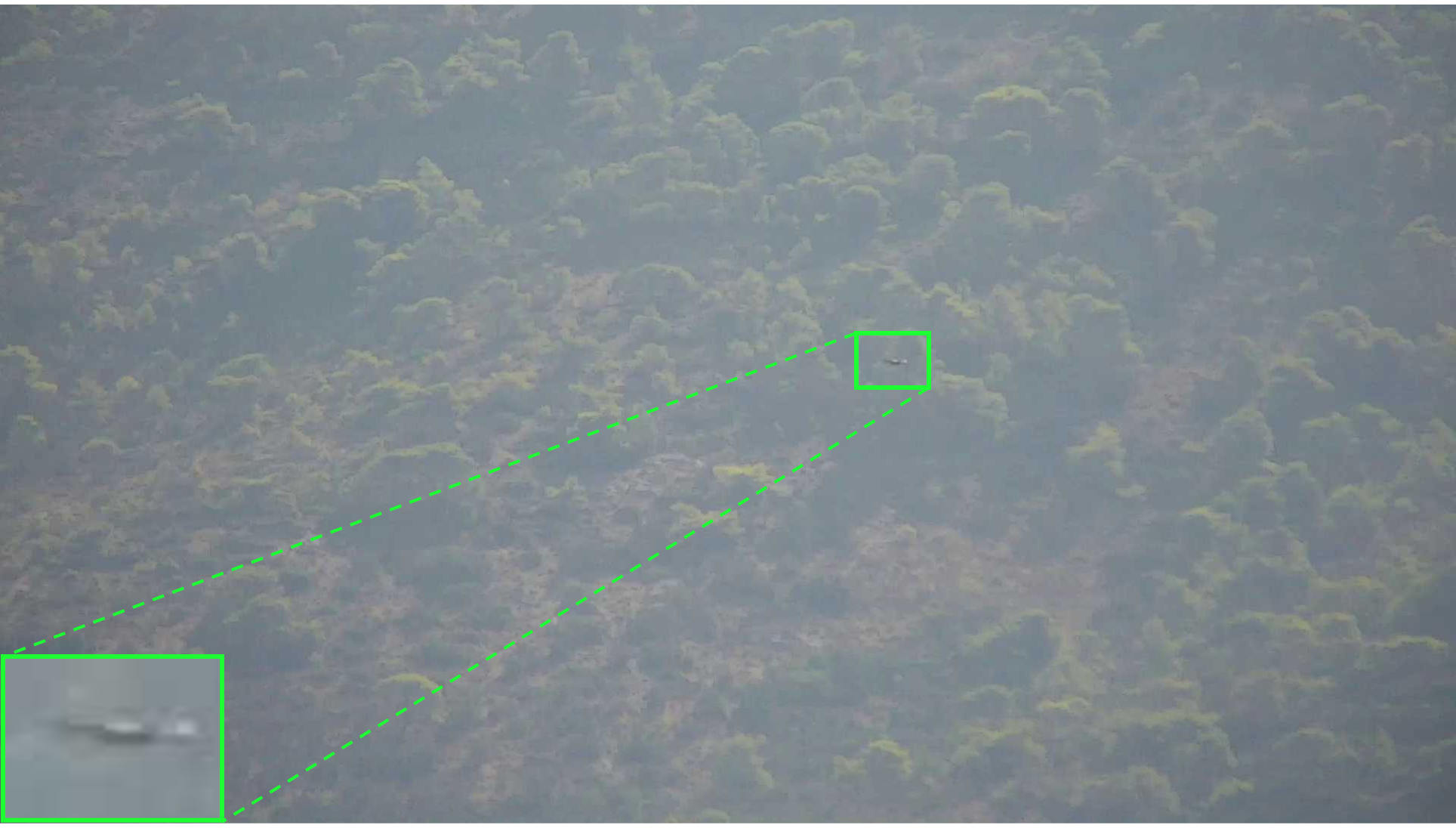}
	\end{minipage}
	\hfill
	\begin{minipage}[b]{0.235\textwidth}
		\centering
		\textbf{YOLO11m-P2}\\[0.5mm]
		\includegraphics[width=\linewidth]{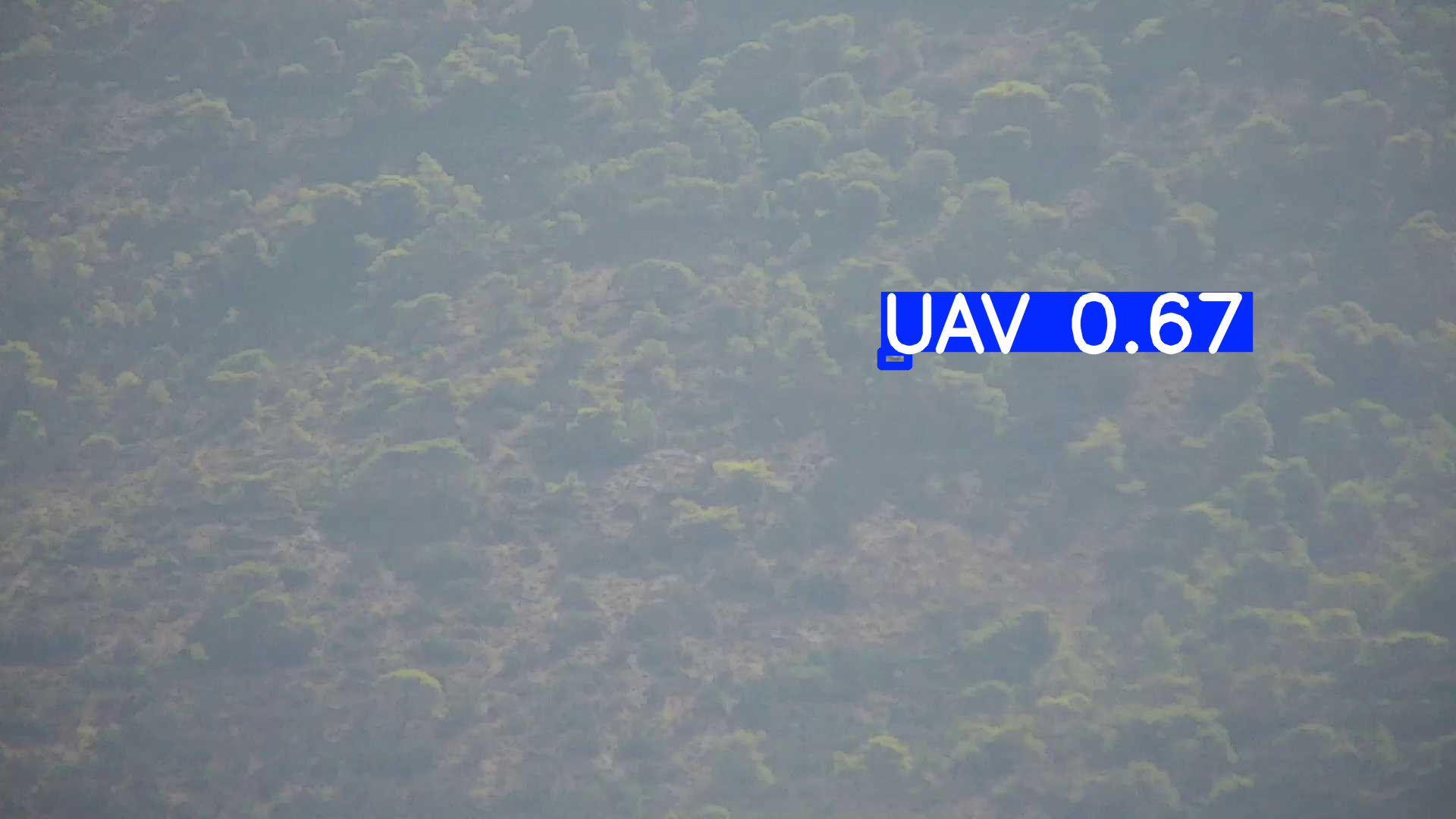}
	\end{minipage}
	\hfill
	\begin{minipage}[b]{0.235\textwidth}
		\centering
		\textbf{CA-YOLO11n}\\[0.5mm]
		\includegraphics[width=\linewidth]{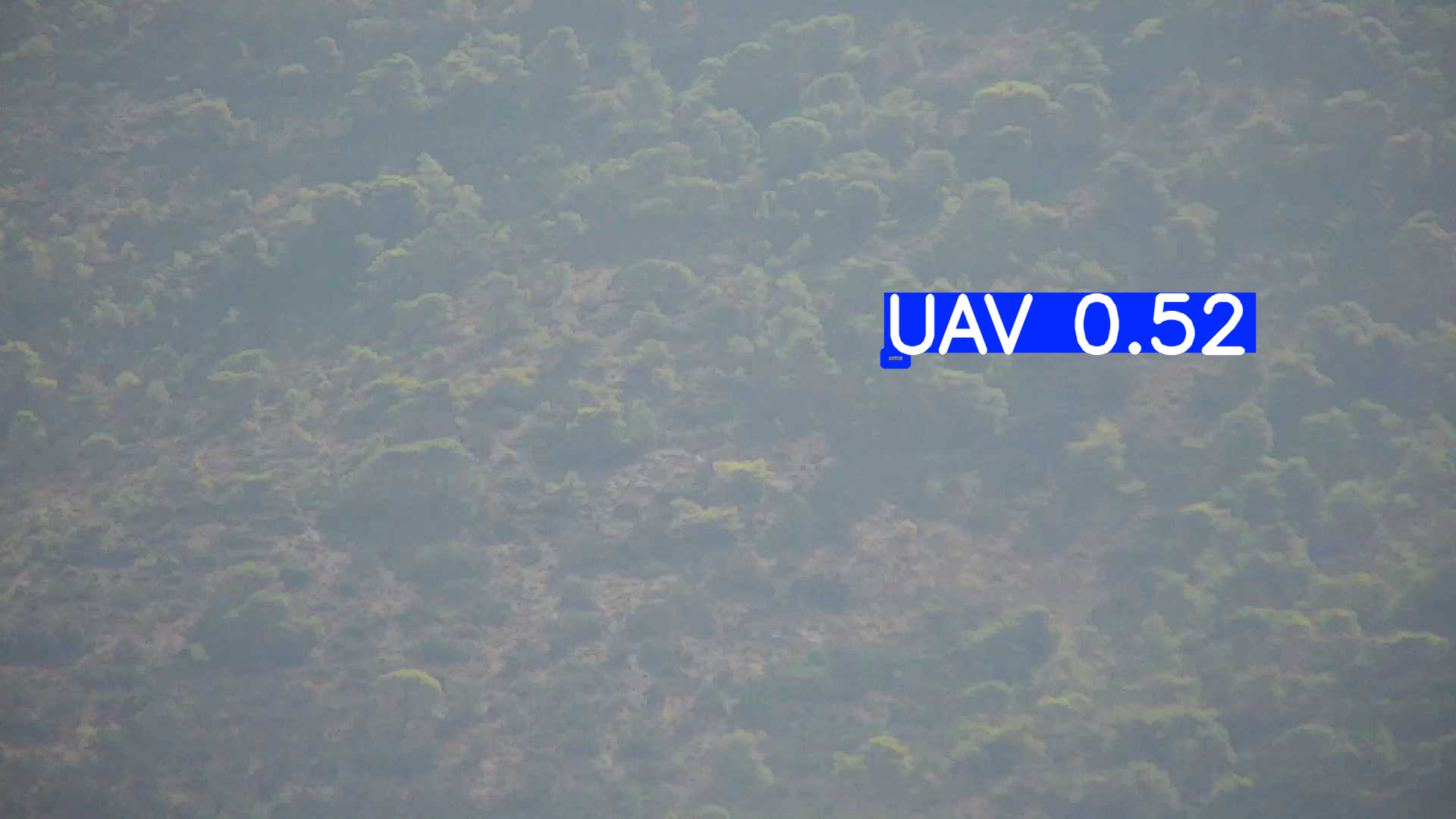}
	\end{minipage}
	\hfill
	\begin{minipage}[b]{0.235\textwidth}
		\centering
		\textbf{YOLO11n-CSCWD}\\[0.5mm]
		\includegraphics[width=\linewidth]{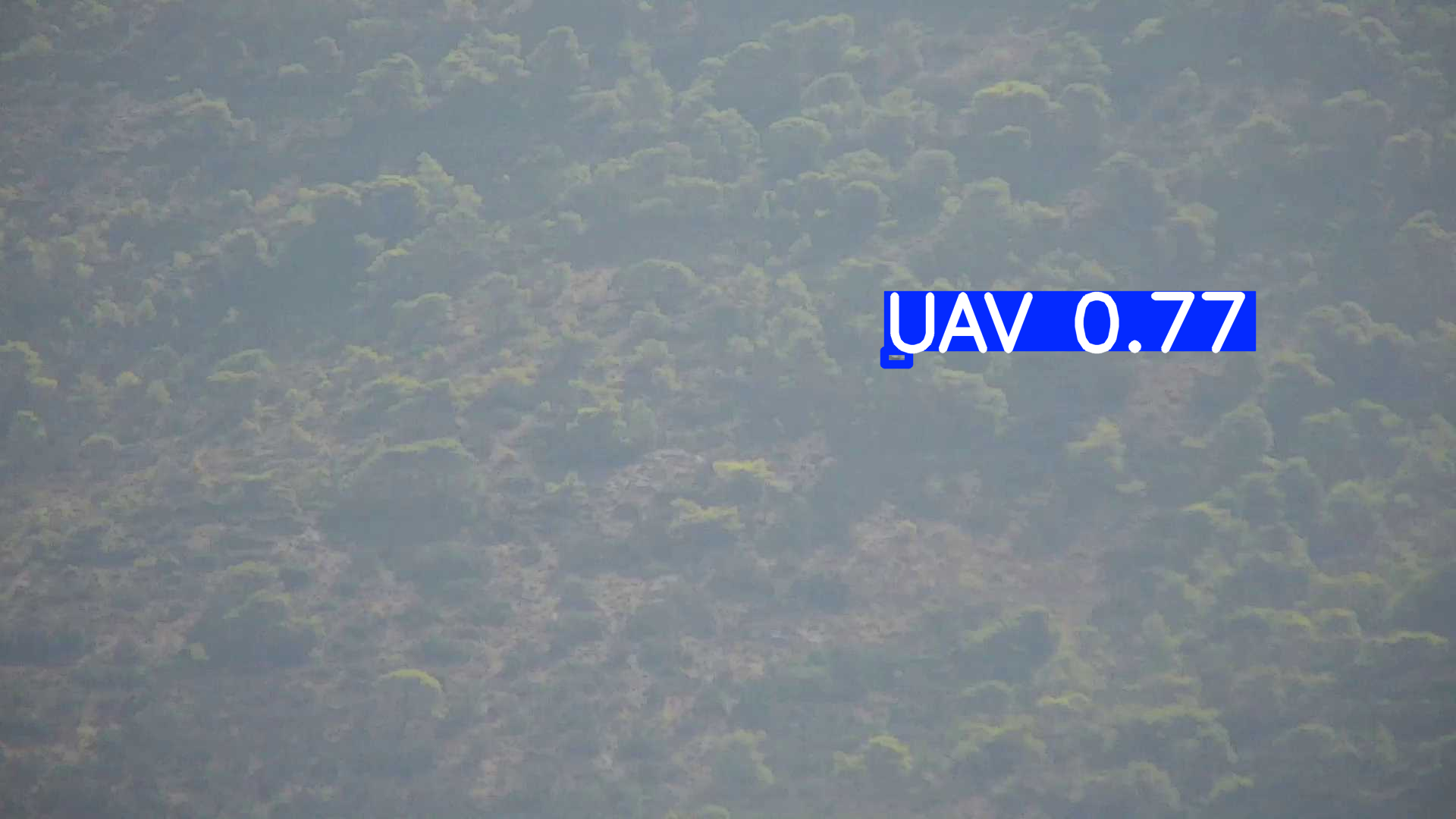}
	\end{minipage}
	
	\par\vspace{2.5mm}

	\begin{minipage}[b]{0.235\textwidth}
		\centering
		\includegraphics[width=\linewidth]{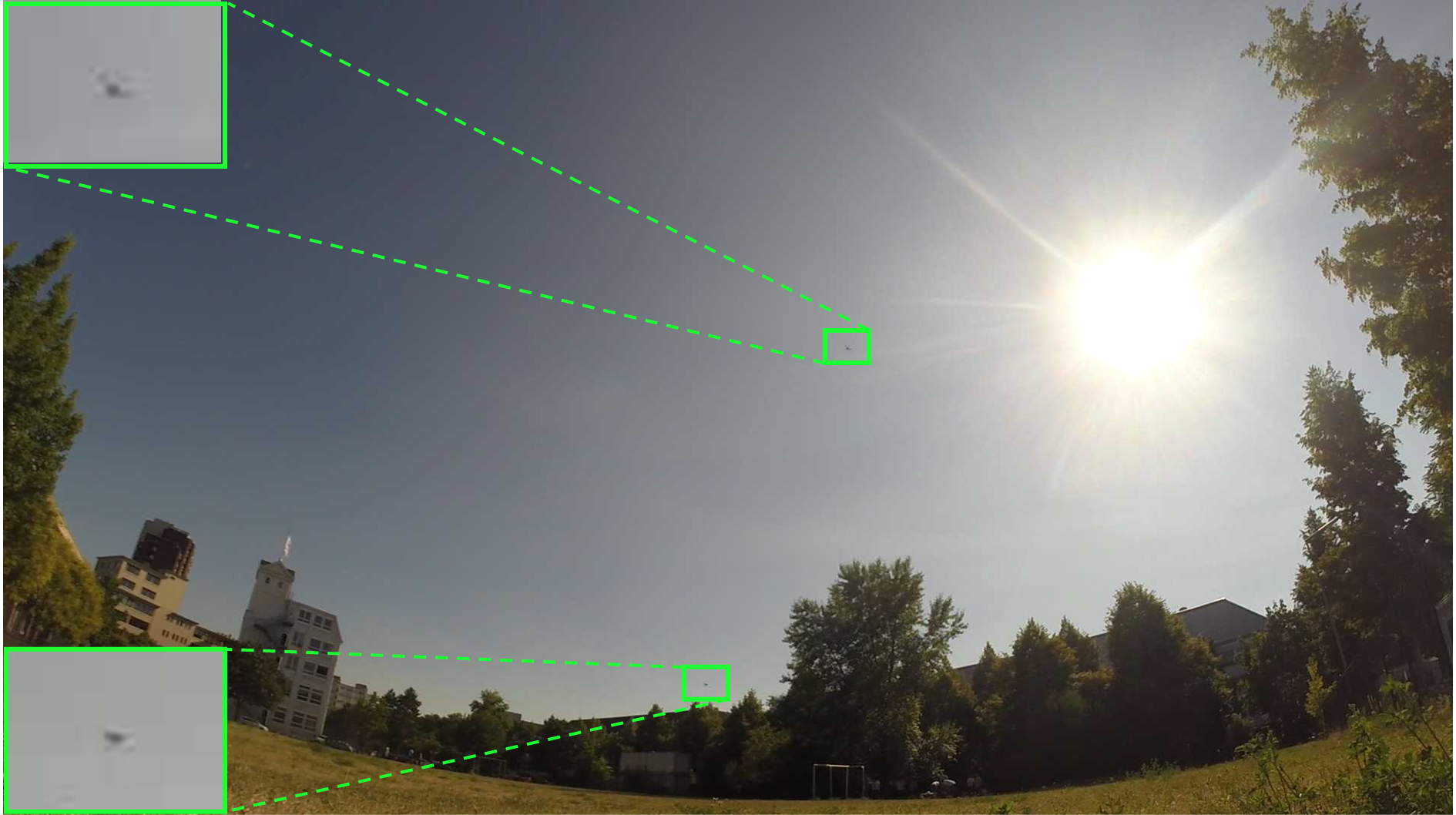}
	\end{minipage}
	\hfill
	\begin{minipage}[b]{0.235\textwidth}
		\centering
		\includegraphics[width=\linewidth]{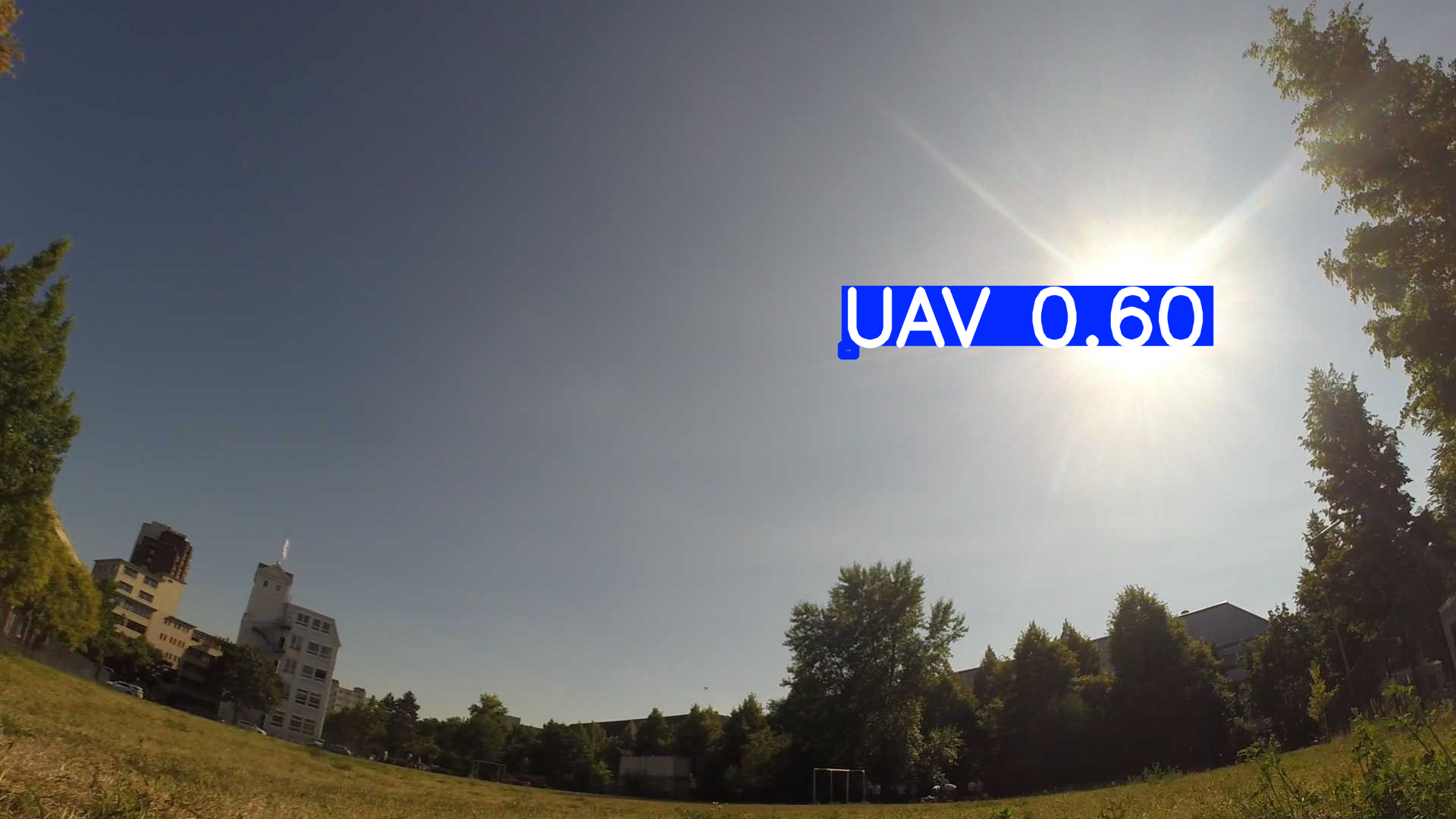}
	\end{minipage}
	\hfill
	\begin{minipage}[b]{0.235\textwidth}
		\centering
		\includegraphics[width=\linewidth]{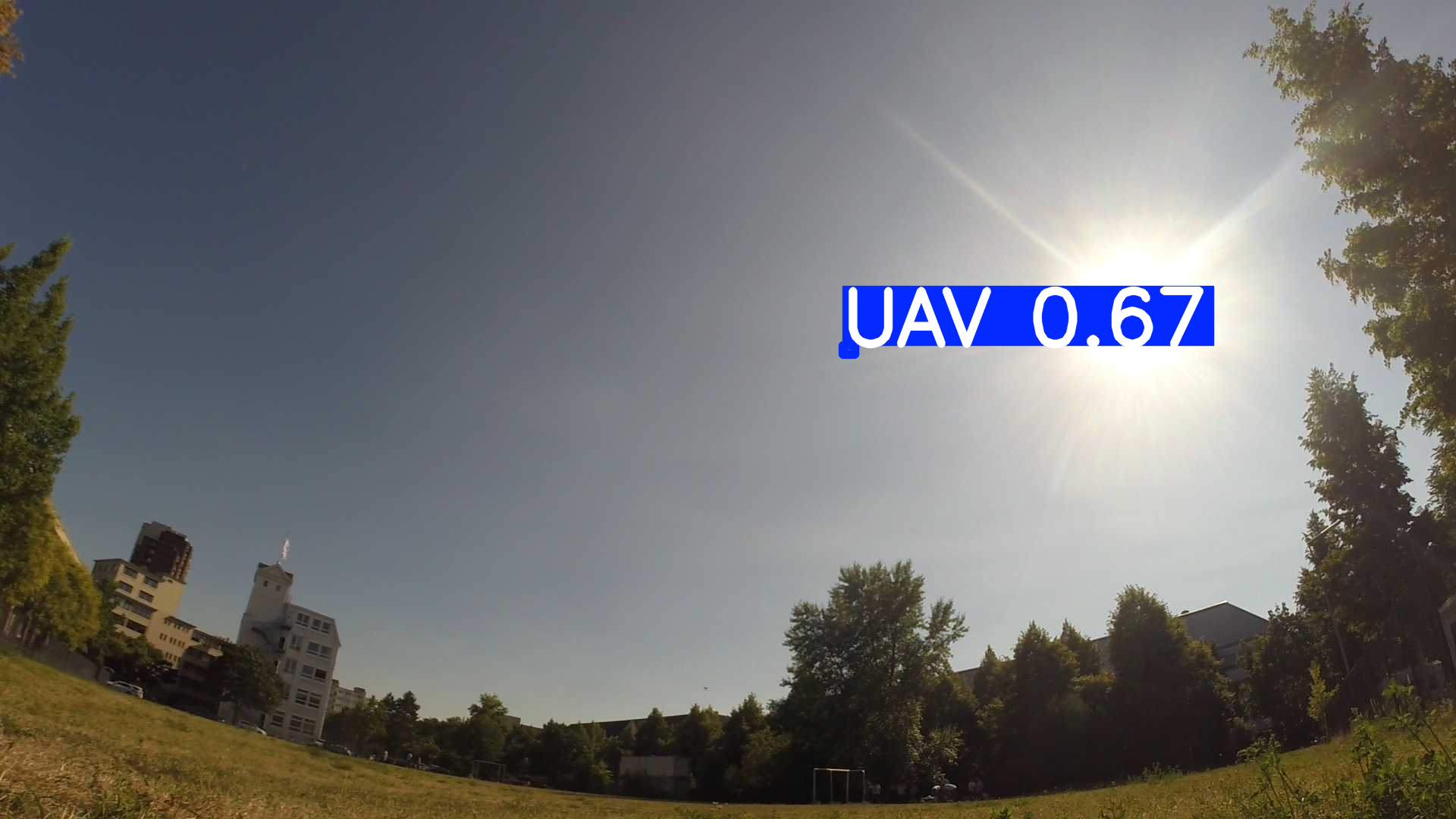}
	\end{minipage}
	\hfill
	\begin{minipage}[b]{0.235\textwidth}
		\centering
		\includegraphics[width=\linewidth]{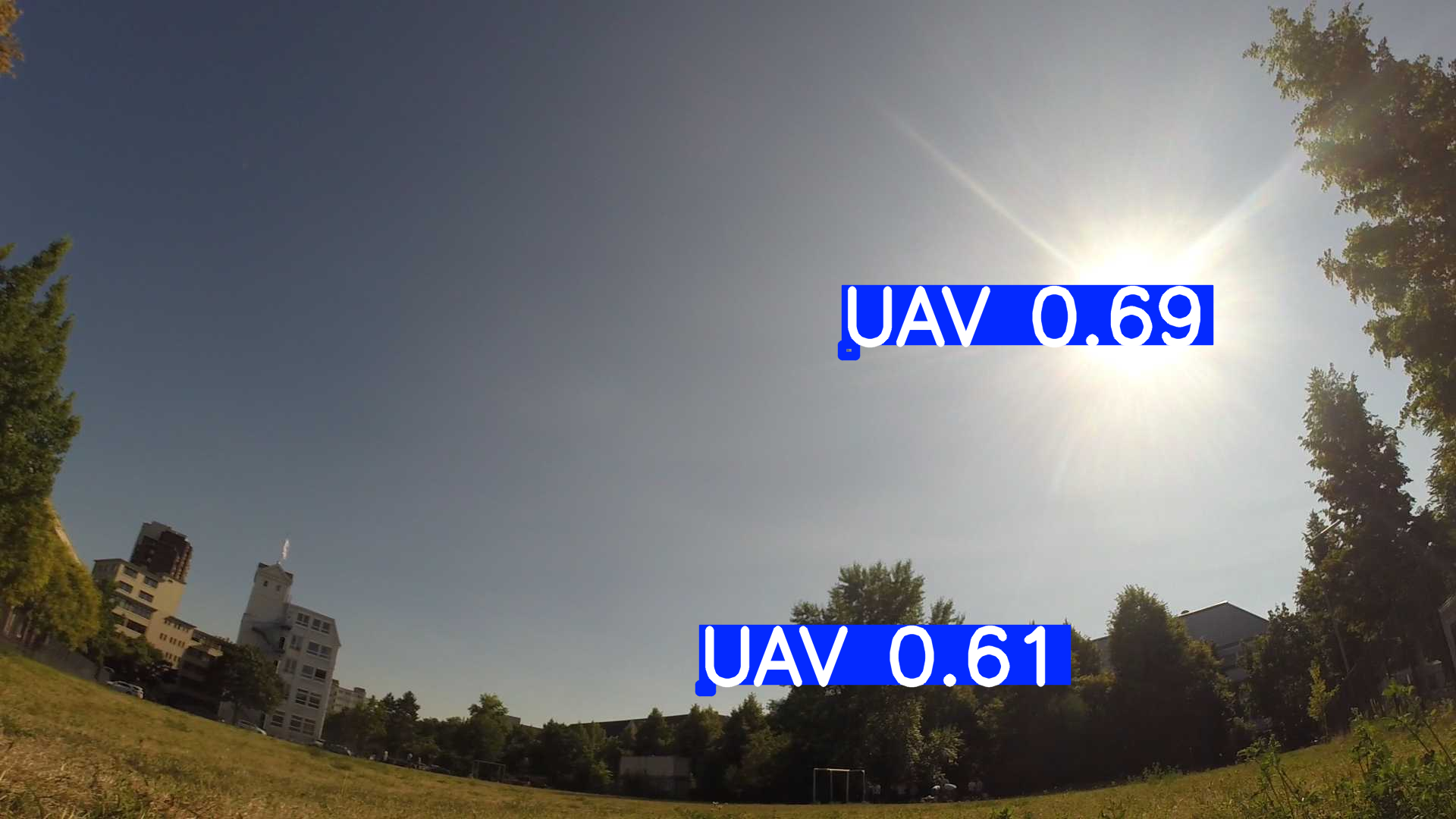}
	\end{minipage}
	
	\caption{
		Qualitative Drone-vs-Bird examples comparing ground truth,
		YOLO11m-P2, CA-YOLO11n, and YOLO11n-CSCWD. The selected cases
		illustrate successful tiny-target detection under challenging visual conditions
	}
	\label{fig:qualitative_results}
\end{figure*}

Target scale alone does not explain sequence difficulty. Inspire contains
comparatively larger resized targets but remains difficult for the controlled
detectors, whereas GoPro~002 combines extremely small targets with the clearest
CSCWD failure case. The reduced qualitative comparison in
Fig.~\ref{fig:qualitative_results} therefore illustrates successful detections
under two challenging visual conditions.

Zero-shot evaluation on DUT-Anti-UAV is reported in
Table~\ref{tab:generalization_dut}. The Drone-vs-Bird-trained checkpoints are
applied directly to the DUT test split without target-domain fine-tuning.
YOLO11n-CSCWD improves CA-YOLO11n by 3.98 points in precision, 1.61 points
in recall, and 1.77 points in mAP@0.5, whereas mAP@0.5:0.95 remains
essentially unchanged (24.00\% versus 24.04\%).

\begin{table}[b]
	\centering
	\caption{Zero-shot DUT-Anti-UAV evaluation without target-domain fine-tuning.}
	\label{tab:generalization_dut}
	
	\footnotesize
	\setlength{\tabcolsep}{1.8pt}
	\renewcommand{\arraystretch}{1.00}
	
	\begin{tabularx}{\columnwidth}{
			@{}
			>{\raggedright\arraybackslash}X
			c c c c
			@{}
		}
		\toprule
		
		\textbf{Model}
		& \textbf{Precision}
		& \textbf{Recall}
		& \textbf{\makecell{mAP\\@0.5}}
		& \textbf{\makecell{mAP\\@0.5:0.95}} \\
		\midrule
		
		CA-YOLO11n
		& 70.61
		& 46.59
		& 48.29
		& \textbf{24.04} \\
		
		YOLO11n-CSCWD
		& \textbf{74.59}
		& \textbf{48.20}
		& \textbf{50.06}
		& 24.00 \\
		
		YOLO11m-P2
		& 83.33
		& 56.34
		& 62.57
		& 30.43 \\
		
		\bottomrule
	\end{tabularx}

\end{table}

\subsection{Ablation Analysis}
\label{sec:ablation}

Two controlled ablations isolate the effects of the feature-distillation
objective and teacher--student feature pairing. The first compares alternative
feature-transfer objectives, whereas the second separates teacher design from
cross-scale alignment. Both ablations retain the same student initialization and data pipeline unless
explicitly stated otherwise.

\subsubsection{Distillation Objective}
\label{sec:distillation_loss_ablation}

CWD~\cite{shu_channel-wise_2021} is compared with Masked Generative
Distillation (MGD)~\cite{avidan_masked_2022} and a hybrid objective combining
$\mathcal{L}_{\mathrm{YOLO}}$, $\mathcal{L}_{\mathrm{CWD}}$, and
$\mathcal{L}_{\mathrm{MGD}}$.

\begin{table}[t]
	\centering
	\caption{Ablation of feature-distillation objectives under the unified seven-sequence validation protocol.}
	\label{tab:distillation_ablation}
	
	\footnotesize
	\renewcommand{\arraystretch}{1.03}
	\setlength{\tabcolsep}{1.4pt}
	
	\begin{tabularx}{\columnwidth}{
			@{} >{\raggedright\arraybackslash}X c c c c @{}}
		\toprule
		
		\textbf{Model}
		& \textbf{Precision}
		& \textbf{Recall}
		& \makecell{\textbf{mAP}\\\textbf{@0.5}}
		& \makecell{\textbf{mAP}\\\textbf{@0.5:0.95}} \\
		\midrule
		
		CA-YOLO11n~\cite{zamani_optimizing_2026}
		& 61.72
		& 56.18
		& 47.25
		& 16.92 \\
		
		YOLO11n-MGD~\cite{avidan_masked_2022}
		& \textbf{63.96}
		& 58.71
		& \textbf{50.53}
		& \textbf{18.00} \\
		
		YOLO11n-Hybrid
		& 63.68
		& 57.16
		& 49.70
		& 17.36 \\
		
		YOLO11n-CSCWD
		& 63.42
		& \textbf{59.73}
		& 50.17
		& 17.30 \\
		
		\bottomrule
	\end{tabularx}
\end{table}

Among the evaluated objectives, MGD achieves the highest mAP@0.5
(50.53\%) and mAP@0.5:0.95 (18.00\%), whereas YOLO11n-CSCWD provides
the highest Recall (59.73\%) with 50.17\% mAP@0.5
(Table~\ref{tab:distillation_ablation}). The 0.36-point mAP@0.5 difference
does not support a claim of universal CWD superiority; the principal CSCWD
contribution is instead the cross-scale feature pairing examined next.

\subsubsection{Teacher Design and Cross-Scale Alignment}
\label{sec:cross_scale_ablation}

The second ablation distinguishes the contribution of the $P_2$-enhanced
teacher from that of cross-scale teacher--student pairing.

\begin{table*}[t]
	\centering
	\caption{Effect of teacher design and feature pairing under the unified
		seven-sequence validation protocol.}
	\label{tab:teacher_ablation}
	
	\footnotesize
	\setlength{\tabcolsep}{4pt}
	\renewcommand{\arraystretch}{1.00}
	
	\begin{tabularx}{\textwidth}{
			@{}
			>{\raggedright\arraybackslash}X
			c c c c
			@{}
		}
		\toprule
		
		\textbf{Configuration}
		& \textbf{Precision}
		& \textbf{Recall}
		& \textbf{mAP@0.5}
		& \textbf{mAP@0.5:0.95} \\
		\midrule
		
		YOLO11m
		& 67.66
		& 60.52
		& 52.84
		& 19.17 \\
		
		YOLO11m-P2
		& \textbf{69.43}
		& \textbf{62.73}
		& \textbf{56.75}
		& \textbf{21.26} \\
			
		\midrule
		
		YOLO11n-CWD (YOLO11m teacher)
		& 56.21
		& 52.45
		& 44.08
		& 15.58 \\
		
		YOLO11n-CWD (YOLO11m-P2, same-scale)
		& \textbf{64.84}
		& 55.36
		& 48.08
		& 16.64 \\
		
		YOLO11n-CSCWD
		& 63.42
		& \textbf{59.73}
		& \textbf{50.17}
		& \textbf{17.30} \\
		
		\bottomrule
	\end{tabularx}
\end{table*}

Adding $P_2$ increases teacher mAP@0.5 from 52.84\% to 56.75\%, whereas
CWD using the teacher without $P_2$ yields only 44.08\% for the student.
With the $P_2$-enhanced teacher, same-scale CWD reaches 48.08\%, while
YOLO11n-CSCWD reaches 50.17\% and increases Recall from 55.36\% to
59.73\%, despite a modest reduction in Precision
(Table~\ref{tab:teacher_ablation}). The 2.09-point mAP@0.5 gain over
same-scale CWD indicates that the student improvement is not explained by
teacher capacity alone and also depends on the proposed cross-scale pairing.

\subsection{Embedded Deployment on Raspberry Pi~5}
\label{sec:edge_deployment}

Raspberry Pi~5 results in Table~\ref{tab:edge_performance} report NCNN-FP16
accuracy and application-level runtime at 640 and 960, while 1280 remains
an accuracy-only setting.

\begin{table*}[t]
	\centering
	\caption{Raspberry Pi~5 deployment results for NCNN-FP16 artifacts.
		Accuracy follows the unified seven-sequence validation protocol. Runtime
		statistics at 640 and 960 are averaged over three complete validation runs;
		1280 reports accuracy only.}
	\label{tab:edge_performance}
	
	\footnotesize
	\renewcommand{\arraystretch}{1.04}
	
	\begin{tabular*}{\textwidth}{
			@{\extracolsep{\fill}}
			l c c c c c c c
			@{}
		}
		\toprule
		
		\textbf{Model}
		& \textbf{\makecell{Input\\size}}
		& \textbf{\makecell{mAP\\@0.5}}
		& \textbf{\makecell{mAP\\@0.5:0.95}}
		& \textbf{\makecell{Mean wall\\latency (ms)}}
		& \textbf{\makecell{P95\\latency (ms)}}
		& \textbf{FPS}
		& \textbf{\makecell{Peak RSS\\(MB)}}
		\\
		\midrule
		
		\multirow{3}{*}{CA-YOLO11n}
		& 640
		& 47.81
		& 16.92
		& 82.29
		& 109.48
		& 12.15
		& 561.23 \\
		
		& 960
		& 47.99
		& 17.54
		& 183.64
		& 213.21
		& 5.45
		& 628.02 \\
		
		& 1280
		& 39.17
		& 13.61
		& --
		& --
		& --
		& -- \\
		
		\midrule
		
		\multirow{3}{*}{YOLO11n-CSCWD}
		& 640
		& \textbf{50.32}
		& \textbf{17.36}
		& 82.32
		& 109.97
		& 12.15
		& 559.81 \\
		
		& 960
		& \textbf{52.85}
		& \textbf{19.38}
		& 183.44
		& 212.05
		& 5.45
		& 624.31 \\
		
		& 1280
		& \textbf{48.53}
		& \textbf{17.76}
		& --
		& --
		& --
		& -- \\
		
		\bottomrule
	\end{tabular*}
\end{table*}

At $640\times640$, YOLO11n-CSCWD raises deployed mAP@0.5 from 47.81\%
to 50.32\% while both models remain at approximately 82.3~ms mean
wall-clock latency and 12.15~FPS. At $960\times960$, the mAP@0.5 advantage
widens to 4.86 points (52.85\% versus 47.99\%) while latency and throughput
again remain essentially unchanged. Peak RSS is also comparable at both
resolutions, consistent with the unchanged student inference graph.

Accuracy is non-monotonic with input resolution. YOLO11n-CSCWD increases from
50.32\% mAP@0.5 at 640 to 52.85\% at 960, but decreases to 48.53\% at
1280; CA-YOLO11n shows a stronger decline. Desktop PyTorch evaluation shows
the same qualitative trend at 1280 (48.17\% for YOLO11n-CSCWD and 39.38\%
for CA-YOLO11n), making a purely NCNN-export-specific explanation unlikely.
Across the measured 640 and 960 operating points, YOLO11n-CSCWD therefore
improves deployed detection accuracy while retaining essentially the same
latency, throughput, and memory footprint as CA-YOLO11n.

\section{Discussion}
\label{sec:discussion}

The results indicate that CSCWD benefits primarily from cross-scale access to
the teacher's high-resolution representation rather than from teacher capacity
or detector scaling alone.

\subsection{Cross-Scale Transfer and Feature Selectivity}
\label{sec:disc_feature_selectivity}

The ablation evidence shows that a stronger teacher is not sufficient for
effective student transfer. Although the $P_2$ branch improves the teacher,
same-scale CWD with YOLO11m-P2 reaches 48.08\% mAP@0.5 for the student,
whereas the proposed teacher-$P_2$ to student-$P_3$ pairing reaches 50.17\%.
The resulting 2.09-point gain, together with the Recall increase from 55.36\%
to 59.73\%, supports the importance of the spatial source and pairing of the
transferred representation. Model scaling alone does not explain this behavior:
the larger YOLO11s reference remains 4.00 points below YOLO11n-CSCWD in
mAP@0.5. The loss ablation provides an important qualification. MGD gives
slightly higher mAP@0.5 and mAP@0.5:0.95 than CSCWD (50.53\% versus
50.17\% and 18.00\% versus 17.30\%), while YOLO11n-CSCWD provides the
highest Recall. The contribution is therefore attributed to cross-scale feature
pairing rather than to universal superiority of CWD. This distinction is
particularly relevant to tiny-object detection, where the location of
high-resolution supervision can matter as much as the choice of transfer loss.

\begin{figure}[!t]
	\centering
	
	\begin{minipage}[b]{0.45\columnwidth}
		\centering
		\textbf{Ground Truth}\\[0.5mm]
		\includegraphics[width=\linewidth]{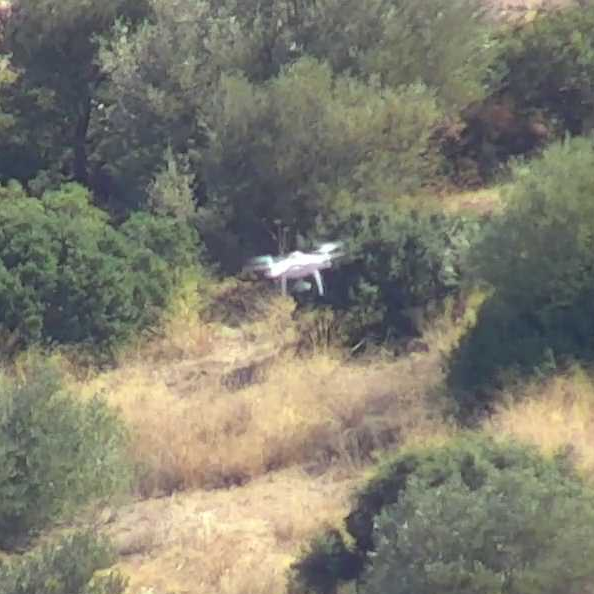}
	\end{minipage}
	\hfill
	\begin{minipage}[b]{0.45\columnwidth}
		\centering
		\textbf{YOLO11m-P2}\\[0.5mm]
		\includegraphics[width=\linewidth]{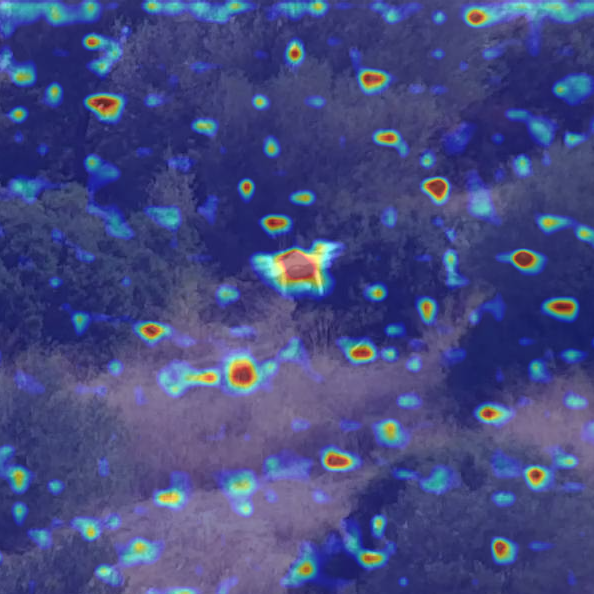}
	\end{minipage}
	
	\vspace{0.5mm}
	
	\begin{minipage}[b]{0.45\columnwidth}
		\centering
		\textbf{CA-YOLO11n}\\[0.5mm]
		\includegraphics[width=\linewidth]{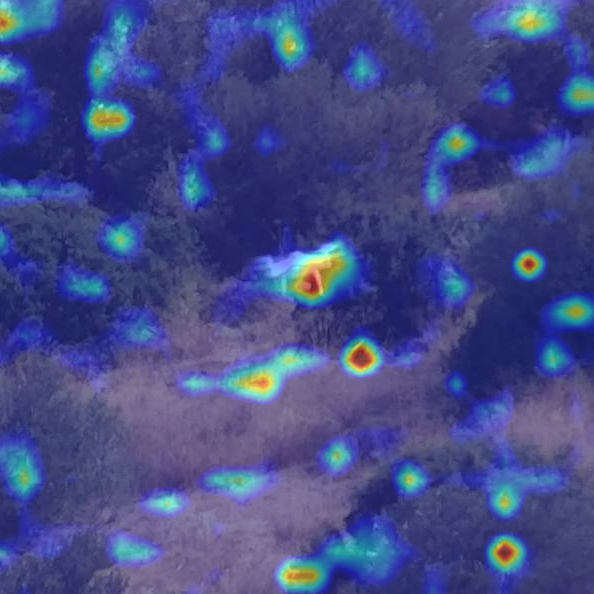}
	\end{minipage}
	\hfill
	\begin{minipage}[b]{0.45\columnwidth}
		\centering
		\textbf{YOLO11n-CSCWD}\\[0.5mm]
		\includegraphics[width=\linewidth]{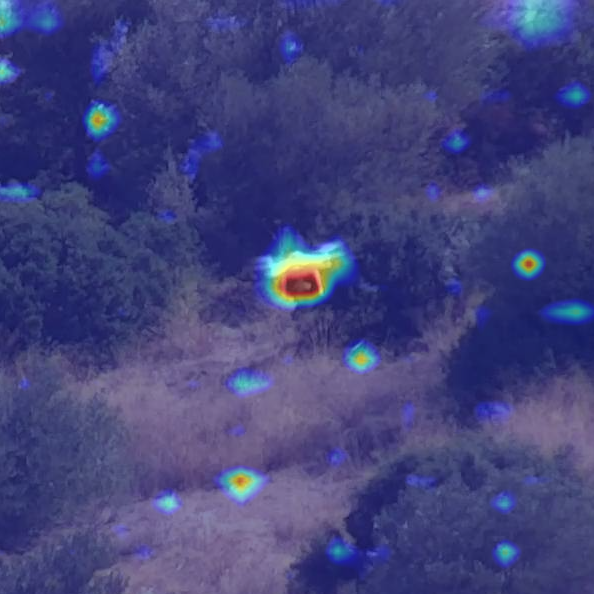}
	\end{minipage}
	
	\caption{
		Target-centered feature responses on a representative Hillside frame.
		YOLO11n-CSCWD shows a more concentrated $P_3$ response than
		CA-YOLO11n; YOLO11m-P2 provides a higher-resolution comparison
	}
	\label{fig:heatmap_cwd}
\end{figure}

The nTCR analysis provides complementary evidence at the feature level.
Across the 11,128 paired frames, YOLO11n-CSCWD has higher nTCR than
CA-YOLO11n on 62.7\% of frames; a paired Wilcoxon signed-rank test gives
$p<0.001$. The effect is modest and sequence dependent, so nTCR is
interpreted only as a diagnostic of intermediate spatial selectivity. This
distinction is important because GoPro~002 shows higher nTCR for
YOLO11n-CSCWD despite lower sequence-level mAP@0.5. Stronger target-region
concentration is therefore not sufficient for improved final detection, where
localization quality, feature resolution, detection-head behavior, and scene
ambiguity also contribute. Figure~\ref{fig:heatmap_cwd} illustrates the same
pattern qualitatively: YOLO11n-CSCWD produces a more target-focused $P_3$
response than CA-YOLO11n in the selected Hillside example, while the teacher $P_2$ response is shown only for higher-resolution comparison.

\subsection{Resolution and Edge-Deployment Trade-offs}
\label{sec:disc_edge_tradeoff}

The Raspberry Pi~5 results show that input resolution acts as an empirical
deployment operating point rather than a monotonic accuracy control. For
YOLO11n-CSCWD, $640\times640$ provides the higher-throughput setting
(50.32\% mAP@0.5 at 12.15~FPS), whereas $960\times960$ provides the highest
measured deployed accuracy (52.85\% at 5.45~FPS). Accuracy then decreases to
48.53\% at $1280\times1280$. CA-YOLO11n exhibits the same qualitative
high-resolution degradation, and desktop PyTorch reproduces the trend,
making a purely NCNN-specific export artifact unlikely to be the sole cause.
Because both detectors were trained at 640, train--test scale mismatch may
contribute, although the experiment does not isolate this mechanism. At a
fixed resolution, the two lightweight models retain essentially identical
latency, throughput, and memory usage, consistent with the unchanged YOLO11n
inference graph after distillation. The resulting trade-off favors 640 when
throughput is prioritized and 960 when the highest measured accuracy is more
important.

\begin{figure}[t]
	\centering
	\includegraphics[width=\columnwidth]{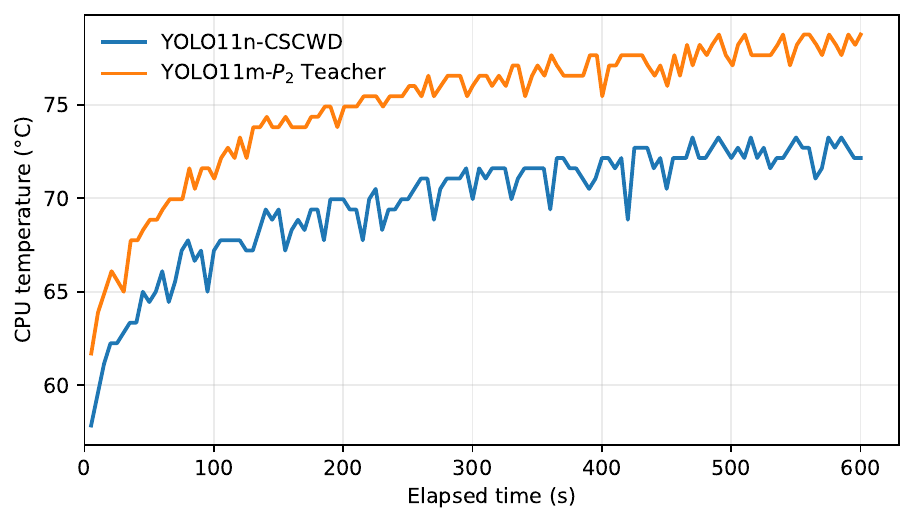}
	\caption{
		Raspberry Pi~5 CPU temperature during 10~min of sustained NCNN-FP16
		inference at $640\times640$. YOLO11n-CSCWD remains cooler than
		YOLO11m-P2, with no observed throttling
	}
	\label{fig:thermal}
\end{figure}

The sustained-load comparison provides a secondary hardware perspective.
YOLO11n-CSCWD sustains 11.85~FPS with a final-120-s mean CPU temperature of
72.49$^{\circ}$C, whereas YOLO11m-P2 reaches 0.94~FPS and 78.13$^{\circ}$C;
no throttling is observed in either run (Fig.~\ref{fig:thermal}). This
difference reflects deployed model scale and computational demand rather than
a direct thermal effect of the distillation loss. Because one controlled
10-min run is used per model, the result is descriptive and should not be
interpreted as a statistical estimate of long-term thermal reliability.

\subsection{Limitations and Future Directions}
\label{sec:disc_limitations}

Zero-shot DUT-Anti-UAV evaluation retains an mAP@0.5 advantage for
YOLO11n-CSCWD, while mAP@0.5:0.95 remains essentially unchanged. The
benefit under domain shift is therefore stronger for detection confidence and
sensitivity than for stricter localization quality, while a substantial gap to
the higher-capacity teacher remains. This result provides cross-dataset
evidence for the proposed approach but does not establish broad generalization
across unseen tiny-object settings. Evaluation remains concentrated on aerial
imagery, with UAVs serving as representative tiny targets. Resolution
sensitivity is measured with models trained at 640 rather than with
resolution-aware retraining. nTCR is correlational and does not establish a
causal link between feature concentration and detection accuracy, while the
thermal comparison is based on one sustained-load run per model. Future work
should examine resolution-aware training, broader tiny-object benchmarks,
hardware-aware compression, and lightweight temporal modeling for transient
misses.

\section{Conclusion}
\label{sec:conclusion}

This work introduced Cross-Scale Channel-wise Knowledge Distillation (CSCWD)
for lightweight tiny-object detection under resource-constrained deployment.
CSCWD transfers the teacher's high-resolution $P_2$ representation to student
$P_3$ during training while preserving the deployed YOLO11n inference graph.
Using aerial UAV imagery as a representative tiny-object setting, YOLO11n-CSCWD
achieves 50.17\% mAP@0.5 under the unified seven-sequence Drone-vs-Bird
validation protocol, improving CA-YOLO11n by 2.92 percentage points.
Cross-scale $P_2$-to-$P_3$ alignment further improves mAP@0.5 by 2.09 points
over same-scale CWD, supporting the importance of teacher--student feature
pairing beyond teacher capacity alone. Zero-shot DUT-Anti-UAV evaluation adds
1.77 mAP@0.5 points without target-domain fine-tuning. On Raspberry Pi~5, the
NCNN-FP16 artifact achieves 50.32\% mAP@0.5 at 82.3\,ms mean wall-clock
latency and 12.15~FPS, with runtime essentially unchanged relative to
CA-YOLO11n. This separation directly preserves the intended accuracy--efficiency
trade-off on embedded edge hardware. These findings support cross-scale
distillation as a practical training-time strategy for improving compact
tiny-object detectors without carrying teacher-side complexity into deployment.

\section*{Statements and Declarations}

\noindent\textbf{Funding.}
No funding was received for this study.

\noindent\textbf{Competing interests.}
The authors have no relevant financial or non-financial interests to disclose.

\noindent\textbf{Author contributions.}
Amir Zamani: Conceptualization, Methodology, Software, Validation,
Formal analysis, Investigation, Data curation, Visualization, and
Writing--original draft.
Zeinab Ghasemi-Naraghi: Supervision, Methodology, Project administration,
and Writing--review and editing.
Both authors reviewed and approved the final manuscript.

\noindent\textbf{Data availability.}
The Drone-vs-Bird dataset is available from the official challenge organizers
upon request and subject to its access conditions. DUT-Anti-UAV and VisDrone
are publicly available from their official repositories. Derived evaluation
outputs and deployment benchmark logs are available from the corresponding
author upon reasonable request, subject to the original dataset licensing
conditions.

\noindent\textbf{Code availability.}
The source code and configuration files are not publicly available
at the time of this preprint release.

\bibliography{references}

@inproceedings{avidan_masked_2022,
  author    = {Yang, Zhendong and Li, Zhe and Shao, Mingqi and Shi, Dachuan and Yuan, Zehuan and Yuan, Chun},
  title     = {Masked Generative Distillation},
  booktitle = {Proc. Eur. Conf. Comput. Vis. (ECCV)},
  pages     = {53--69},
  year      = {2022},
  doi       = {10.1007/978-3-031-20083-0_4}
}

@article{coluccia_drone-vs-bird_2024,
  author  = {Coluccia, Angelo and Fascista, Alessio and Sommer, Lars and Schumann, Arne and Dimou, Anastasios and Zarpalas, Dimitrios},
  title   = {The Drone-vs-Bird Detection Grand Challenge at ICASSP 2023: A Review of Methods and Results},
  journal = {IEEE Open J. Signal Process.},
  volume  = {5},
  pages   = {766--779},
  year    = {2024},
  doi     = {10.1109/OJSP.2024.3379073}
}

@inproceedings{coluccia_drone-vs-bird_2025,
  author    = {Coluccia, Angelo and Fascista, Alessio and Dimou, Anastasios and Zarpalas, Dimitrios and Sommer, Lars and Schumann, Arne and Mele, Emanuele},
  title     = {The Drone-vs-Bird Detection Grand Challenge at IJCNN 2025},
  booktitle = {Proc. Int. Joint Conf. Neural Netw. (IJCNN)},
  pages     = {1--8},
  year      = {2025},
  doi       = {10.1109/IJCNN64981.2025.11228314}
}

@inproceedings{hinton_distilling_2015,
  author    = {Hinton, Geoffrey and Vinyals, Oriol and Dean, Jeff},
  title     = {Distilling the Knowledge in a Neural Network},
  booktitle = {NIPS Deep Learning and Representation Learning Workshop},
  year      = {2015}
}

@inproceedings{laroca_improving_2025,
  author    = {Laroca, Rayson and Dos Santos, Marcelo and Menotti, David},
  title     = {Improving Small Drone Detection Through Multi-Scale Processing and Data Augmentation},
  booktitle = {Proc. Int. Joint Conf. Neural Netw. (IJCNN)},
  pages     = {1--8},
  year      = {2025},
  doi       = {10.1109/IJCNN64981.2025.11227421}
}

@article{mansourian_comprehensive_2025,
  author  = {Mansourian, Amir M. and Ahmadi, Rozhan and Ghafouri, Masoud and Babaei, Amir Mohammad and Golezani, Elaheh Badali and Ghamchi, Zeynab Yasamani and Ramezanian, Vida and Taherian, Alireza and Dinashi, Kimia and Miri, Amirali and Kasaei, Shohreh},
  title   = {A Comprehensive Survey on Knowledge Distillation},
  journal = {Transactions on Machine Learning Research},
  year    = {2025}
}

@article{mittal_comprehensive_2024,
  author  = {Mittal, Payal},
  title   = {A Comprehensive Survey of Deep Learning-Based Lightweight Object Detection Models for Edge Devices},
  journal = {Artif. Intell. Rev.},
  volume  = {57},
  pages   = {242},
  year    = {2024},
  doi     = {10.1007/s10462-024-10877-1}
}

@article{nikouei_small_2025,
  author  = {Nikouei, Mahya and Baroutian, Bita and Nabavi, Shahabedin and Taraghi, Fateme and Aghaei, Atefe and Sajedi, Ayoob and Moghaddam, Mohsen Ebrahimi},
  title   = {Small Object Detection: A Comprehensive Survey on Challenges, Techniques and Real-World Applications},
  journal = {Intell. Syst. Appl.},
  volume  = {27},
  pages   = {200561},
  year    = {2025},
  doi     = {10.1016/j.iswa.2025.200561}
}

@inproceedings{romero_fitnets_2015,
  author    = {Romero, Adriana and Ballas, Nicolas and Kahou, Samira Ebrahimi and Chassang, Antoine and Gatta, Carlo and Bengio, Yoshua},
  title     = {FitNets: Hints for Thin Deep Nets},
  booktitle = {Int. Conf. Learn. Represent. (ICLR)},
  year      = {2015}
}

@inproceedings{sawada_tiny_2025,
  author    = {Sawada, Azusa and Ogawa, Takuya and Higa, Kyota},
  title     = {Tiny Drone Detection from Videos with Tracking-Based Temporal Lost Compensation},
  booktitle = {Proc. Int. Joint Conf. Neural Netw. (IJCNN)},
  pages     = {1--6},
  year      = {2025},
  doi       = {10.1109/IJCNN64981.2025.11228641}
}

@inproceedings{shu_channel-wise_2021,
  author    = {Shu, Changyong and Liu, Yifan and Gao, Jianfei and Yan, Zheng and Shen, Chunhua},
  title     = {Channel-wise Knowledge Distillation for Dense Prediction},
  booktitle = {Proc. IEEE/CVF Int. Conf. Comput. Vis. (ICCV)},
  pages     = {5291--5300},
  year      = {2021},
  doi       = {10.1109/ICCV48922.2021.00526}
}

@misc{ultralytics_yolov11_2024,
  author       = {Jocher, Glenn and Qiu, Jing},
  title        = {Ultralytics YOLO11},
  year         = {2024},
  howpublished = {Online},
  url          = {https://github.com/ultralytics/ultralytics},
  note         = {Version 11.0.0. Accessed 16 September 2026}
}

@inproceedings{wong_wrn-yolo_2025,
  author    = {Wong, Yi Jie and Voon, Wingates and Tham, Mau-Luen and Kwan, Ban-Hoe and Chang, Yoong Choon and Hum, Yan Chai},
  title     = {WRN-YOLO: An Improved YOLO for Drone Detection using Wide ResNet},
  booktitle = {Proc. Int. Joint Conf. Neural Netw. (IJCNN)},
  pages     = {1--8},
  year      = {2025},
  doi       = {10.1109/IJCNN64981.2025.11229346}
}

@inproceedings{yang_focal_2022,
  author    = {Yang, Zhendong and Li, Zhe and Jiang, Xiaohu and Gong, Yuan and Yuan, Zehuan and Zhao, Danpei and Yuan, Chun},
  title     = {Focal and Global Knowledge Distillation for Detectors},
  booktitle = {Proc. IEEE/CVF Conf. Comput. Vis. Pattern Recognit. (CVPR)},
  pages     = {4633--4642},
  year      = {2022},
  doi       = {10.1109/CVPR52688.2022.00460}
}

@inproceedings{zagoruyko_paying_2017,
  author    = {Zagoruyko, Sergey and Komodakis, Nikos},
  title     = {Paying More Attention to Attention: Improving the Performance of Convolutional Neural Networks via Attention Transfer},
  booktitle = {Int. Conf. Learn. Represent. (ICLR)},
  year      = {2017}
}

@inproceedings{zamani_optimizing_2026,
  author    = {Zamani, Amir and Abedini, Zeinab},
  title     = {Optimizing Data Augmentation for Real-Time Small UAV Detection: A Lightweight Context-Aware Approach},
  booktitle = {34th Int. Conf. Electr. Eng. (ICEE)},
  year      = {2026},
  doi       = {10.48550/arXiv.2604.19999},
  note      = {Accepted for presentation}
}

@article{zhao_vision-based_2022,
  author  = {Zhao, Jie and Zhang, Jingshu and Li, Dongdong and Wang, Dong},
  title   = {Vision-Based Anti-UAV Detection and Tracking},
  journal = {IEEE Trans. Intell. Transp. Syst.},
  volume  = {23},
  pages   = {25323--25334},
  year    = {2022},
  doi     = {10.1109/TITS.2022.3177627}
}

@article{zhou_improved_2025,
  author  = {Zhou, Sicheng and Yang, Lei and Liu, Huiting and Zhou, Chongqin and Liu, Jiacheng and Wang, Yang and Zhao, Shuai and Wang, Keyi},
  title   = {Improved YOLO for Long Range Detection of Small Drones},
  journal = {Sci. Rep.},
  volume  = {15},
  pages   = {12280},
  year    = {2025},
  doi     = {10.1038/s41598-025-95580-z}
}

@article{zhu_detection_2022,
  author  = {Zhu, Pengfei and Wen, Longyin and Du, Dawei and Bian, Xiao and Fan, Heng and Hu, Qinghua and Ling, Haibin},
  title   = {Detection and Tracking Meet Drones Challenge},
  journal = {IEEE Trans. Pattern Anal. Mach. Intell.},
  volume  = {44},
  pages   = {7380--7399},
  year    = {2022},
  doi     = {10.1109/TPAMI.2021.3119563}
}

@inproceedings{zhu_scalekd_2023,
  author    = {Zhu, Yichen and Zhou, Qiqi and Liu, Ning and Xu, Zhiyuan and Ou, Zhicai and Mou, Xiaofeng and Tang, Jian},
  title     = {ScaleKD: Distilling Scale-Aware Knowledge in Small Object Detector},
  booktitle = {Proc. IEEE/CVF Conf. Comput. Vis. Pattern Recognit. (CVPR)},
  pages     = {19723--19733},
  year      = {2023},
  doi       = {10.1109/CVPR52729.2023.01889}
}

@misc{ultralytics_yolov8_2023,
  author       = {Jocher, Glenn and Chaurasia, Ayush and Qiu, Jing},
  title        = {Ultralytics YOLOv8},
  year         = {2023},
  howpublished = {Online},
  url          = {https://github.com/ultralytics/ultralytics},
  note         = {Version 8.0.0. Accessed 16 September 2026}
}

@inproceedings{wang_yolov10_2024,
  author    = {Wang, Ao and Chen, Hui and Liu, Lihao and Chen, Kai
               and Lin, Zijia and Han, Jungong and Ding, Guiguang},
  title     = {{YOLOv10}: Real-Time End-to-End Object Detection},
  booktitle = {Adv. Neural Inf. Process. Syst.},
  volume    = {37},
  pages     = {107984--108011},
  year      = {2024},
  doi       = {10.52202/079017-3429}
}

@misc{sparkfun_raspberrypi5_image,
  author = {{SparkFun Electronics}},
  title  = {Raspberry Pi 5},
  year   = {2023},
  url    = {https://commons.wikimedia.org/wiki/File:23551-raspberry-pi-5.jpg}
}

\end{document}